\documentclass{article}

\usepackage{microtype}
\usepackage{graphicx}
\usepackage{subfigure}
\usepackage{booktabs} 
\usepackage{colortbl} 
\usepackage{xcolor}

\usepackage{hyperref}

\usepackage[accepted]{icml2024}

\usepackage{amsmath}
\usepackage{amssymb}
\usepackage{mathtools}
\usepackage{amsthm}

\usepackage{tcolorbox}

\makeatletter
\def\tcb@proc@counter@auto#1{%
  \newcounter{tcb@cnt@#1}%
  \csxdef{tcb@cnt@#1}{tcb@cnt@#1}%
  \tcb@proc@counter@autoanduse{#1}%
  \ifcsname resetcounteronoverlays\endcsname
  \resetcounteronoverlays{tcb@cnt@#1}
  \fi
}
\makeatother
\newtcolorbox[auto counter]{numberedbox}[2][]{%
colback=lightgray!5,colframe=gray!40!black,center,title=Prompt ~\thetcbcounter: #2,#1, left skip=0pt, right skip=0pt, width=\linewidth}

\usepackage[capitalize,noabbrev]{cleveref}

\theoremstyle{plain}

\theoremstyle{definition}

\theoremstyle{remark}

\usepackage[textsize=tiny]{todonotes}

\icmltitlerunning{Robust Planning with LLM-Modulo Framework: Case Study in Travel Planning }

\begin{document}

\twocolumn[
\icmltitle{Robust Planning with LLM-Modulo Framework: Case Study in Travel Planning }



\icmlsetsymbol{equal}{*}

\begin{icmlauthorlist}
\icmlauthor{Atharva Gundawar}{equal,yyy}
\icmlauthor{Mudit Verma}{equal,yyy}
\icmlauthor{Lin Guan}{yyy}
\icmlauthor{Karthik Valmeekam}{yyy}
\icmlauthor{Siddhant Bhambri}{yyy}
\icmlauthor{Subbarao Kambhampati}{yyy}

\icmlaffiliation{yyy}{School of Computing and AI, ASU}


\end{icmlauthorlist}

\icmlcorrespondingauthor{Atharva Gundawar}{agundawa@asu.edu}
\icmlcorrespondingauthor{Mudit Verma}{muditverma@asu.edu}

\icmlkeywords{Large Language Models, Reasoning and Planning, Travel Planning}

\vskip 0.3in
]



\printAffiliationsAndNotice{\icmlEqualContribution} 

\section{Introduction}
\vspace{-0.2cm}
As the applicability of Large Language Models (LLMs) extends beyond traditional text processing tasks, there is a burgeoning interest in their potential to excel in planning and reasoning assignments, realms traditionally reserved for System 2 cognitive competencies \cite{kahneman2011thinking}. Despite their perceived versatility, the research community is still unraveling effective strategies to harness these models in such complex domains. While there are studies showing LLMs are not able to support robust planning \cite{Verma_2024, stechly2024chain, valmeekam2022large, verma2024theory}, there is some consensus that they can help planning in a more integrated architecture \cite{kambhampati2024llms}. The recent discourse introduced by the paper on LLM Modulo \cite{kambhampati2024llms} marks a significant stride, proposing a conceptual framework that enhances the integration of LLMs into diverse planning and reasoning activities. Of interest to this paper is to realize the LLM Modulo Framework for a Planning problem. As motivated by \cite{xie2024travelplanner} Travel planning remains a complex domain, involving choices on destinations, accommodations, transport, and activities, which necessitates managing long-term dependencies and logical reasoning. This complexity makes travel planning an ideal domain to assess the reasoning abilities of planners. Utilizing the Travel Planning Benchmark \cite{xie2024travelplanner}, we aim to determine if language agents can handle realistic scenarios akin to human operations. Despite advanced techniques like ReAct\cite{yao2022react} and Chain of Thought\cite{wei2022chain}, these models achieve less than 1\% accuracy, compared to humans who score 100\%.

The benchmark provides user queries in nautral langauge and an evaluation methodology for validating solution plans / itineraries obtained via LLM agents. In this paper, we will revisit the various abstract components suggested in the LLM-Modulo framework and realize it for the TravelPlanning domain. In this generate-test planning paradigm, the LLMs play several helpful roles such as the generator (to generate the plan or travel itinerary), reformulator (or translator, for converting natural language queries to structured output parseable by other components) and critic extraction (to implement model based critics responsible for testing the LLM generated plan and backprompting the LLM for fixing known issues). 

While popular methods of enhancing reasoning abilities of LLMs such as Chain of Thought, ReAct, and Reflexion achieve a meager 0\%, 0.6\%, and 0\% with GPT3.5-Turbo\cite{openai2022gpt35} respectively , our operationalization of the LLM-Modulo framework for TravelPlanning domain provides a remarkable improvement, enhancing baseline performances by 4.6x for GPT4-Turbo\cite{openai2023gpt4} and even more for older models like GPT3.5-Turbo from 0\% to 5\%. Furthermore, we highlight the other useful roles of LLMs in the planning pipeline, as suggested in LLM-Modulo, can be reliably operationalized such as extraction of useful critics and reformulator for critics.

\section{Background and Setup}
\subsection{Travel Planning domain}
\vspace{-0.2cm}

\begin{figure*}[ht]
    \centering
    \includegraphics[width=0.75\linewidth]{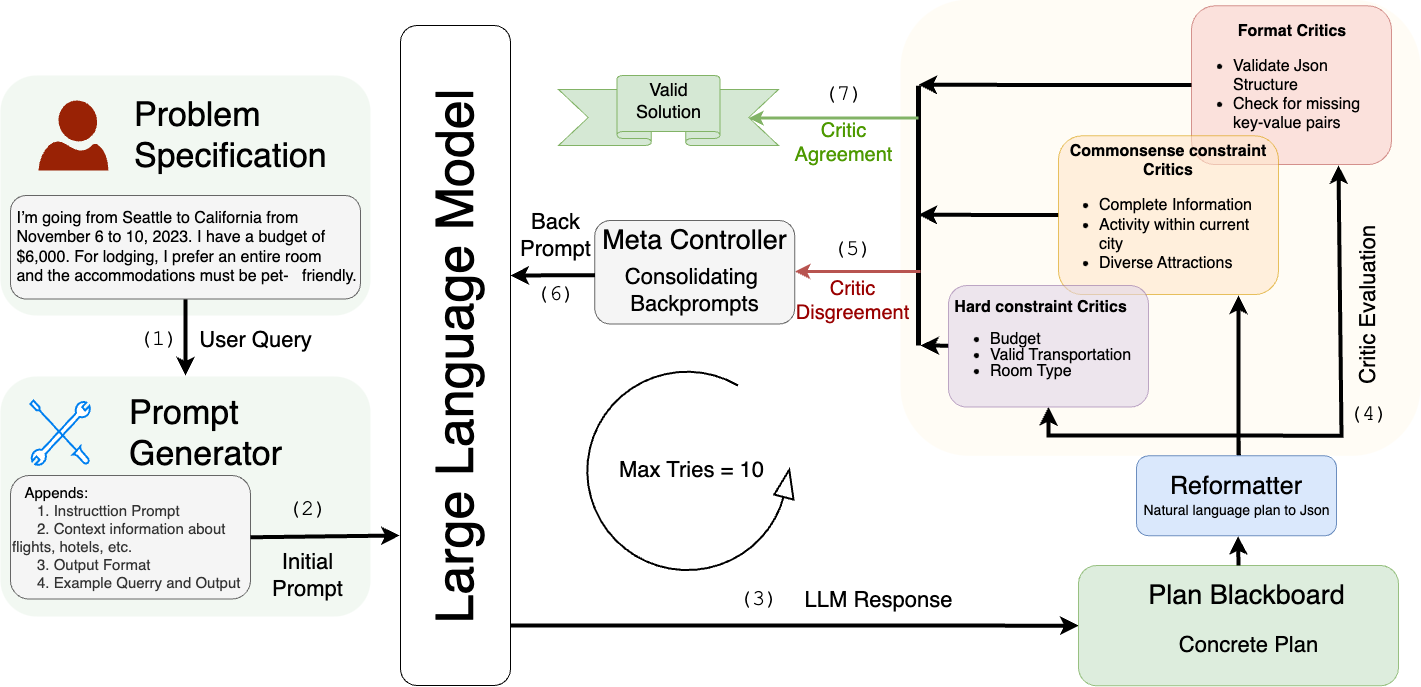}
    \caption{LLM Modulo Framework for Travel Planning}
    \label{fig:llm_modulo_TP}
\end{figure*}

The domain provides a sandboxed environment to generate itineraries for travel plan queries simulated using various datasets such as flights, restaurants, distances between locations, available transport options, accommodation choices to name a few. The TravelPlanning domain evaluates generated plans based on hard constraints and commonsense constraints. We use the recommended dataset (validation dataset) with 180 queries for all our experiments. Use of test set requires an official submission to TravelPlanning leaderboard which is left for future work. Furthermore, as a first investigation of LLM-Modulo for TravelPlanning we restrict our discussion to the \textit{sole-planning} mode which simplifies the objective for the LLMs. In this setting, the domain provides necessary context (that would otherwise be obtained by querying the various datasets) to the LLM instead of expecting the LLM to perform Tool Use \cite{schick2024toolformer,paranjape2023art,hsieh2023tool}. The top performing models in this simplified mode achieves 4.4\% (GPT-4-Turbo) and 0.6\% (across various prompt-engineering tricks with GPT-3.5-Turbo).

Example : A query can be for a 3-day trip from Washington to Myrtle Beach for one traveler with a \$1,400 budget, running from March 13th to 15th, 2022. There are no specific preferences regarding accommodation, cuisine, room type, or transport method. Key attributes captured in each query include the origin city (\texttt{org: Washington}), destination city (\texttt{dest: Myrtle Beach}), trip duration (\texttt{days: 3}), number of cities visited (\texttt{visiting\_city\_number: 1}), number of travelers (\texttt{people\_number: 1}), and no local constraints (\texttt{local\_constraint: none}). The (\textit{budget: \$1,400}) is set to cover all travel expenses, ensuring accurate and efficient travel planning within financial limits.

\subsection{LLM Modulo}
\vspace{-0.2cm}
The \textit{LLM-Modulo framework} introduced by \cite{kambhampati2024llms} establishes a robust iterative interaction between a base generative model, specifically a large language model (LLM), and a suite of external verifiers. These verifiers critically assess the LLM's outputs. Should the output not meet predefined criteria, these external critics provide feedback to the LLM, prompting necessary adjustments. Essentially, the work provides various uses of LLMs in the planning pipeline such as idea-generators, translators, problem specification enrichment, critic/model acquisition to name a few. This work instantiates several abstract roles of LLMs as presented in LLM-Modulo framework specific to the TravelPlanning \cite{xie2024travelplanner} domain.

\section{Instantiating LLM Modulo for Travel Planning}
\vspace{-0.2cm}

\begin{table*}[ht]
    \centering
    \begin{tabular}{lc|cc|cc|c}
        \toprule
        \textbf{Model} & \multicolumn{1}{c}{\textbf{Delivery}} & \multicolumn{2}{c}{\textbf{Commonsense}} & \multicolumn{2}{c}{\textbf{Hard}} & \multicolumn{1}{c}{\textbf{Final Pass}} \\
         & \multicolumn{1}{c}{\textbf{Rate}} & \multicolumn{2}{c}{\textbf{Pass Rate}} & \multicolumn{2}{c}{\textbf{Pass Rate}} & \multicolumn{1}{c}{\textbf{Rate}} \\

        & \textbf{} & \textbf{Micro} & \textbf{Macro} & \textbf{Micro} & \textbf{Macro} & \textbf{} \\
        \midrule
        Direct $_{\rm GPT-3.5-Turbo}$ & 99.4 & 61.5 & 3.9 & 11.2 & 2.8 & 0.0 \\
        Direct $_{\rm GPT-4-Turbo}$ & 100 & 84.9 & 25.6 & 51.9 & 24.4 & 4.4 \\

        \textcolor{gray!75}{CoT$_{\rm GPT-3.5-Turbo}$} & \textcolor{gray!75}{\text{100}} & \textcolor{gray!75}{\text{66.3}} & \textcolor{gray!75}{\text{3.3}} & \textcolor{gray!75}{\text{11.9}} & \textcolor{gray!75}{\text{5}} & \textcolor{gray!75}0 \\
        \textcolor{gray!75}{ReAct$_{\rm GPT-3.5-Turbo}$} & \textcolor{gray!75}{\text{82.2}} & \textcolor{gray!75}{\text{47.6}} & \textcolor{gray!75}{\text{3.9}} & \textcolor{gray!75}{\text{11.4}} & \textcolor{gray!75}{\text{6.7}} & \textcolor{gray!75}{0.6} \\
        \textcolor{gray!75}{Reflexion$_{\rm GPT-3.5-Turbo}$} & \textcolor{gray!75}{\text{93.9}} & \textcolor{gray!75}{\text{53.8}} & \textcolor{gray!75}{\text{2.8}} & \textcolor{gray!75}{\text{11}} & \textcolor{gray!75}{\text{2.8}} & \textcolor{gray!75}{0} \\

        \textbf{LLM Modulo [All]$_{\rm GPT-3.5-Turbo}$} & \textbf{97.8} & \textbf{59.8} & \textbf{13.3} & \textbf{14} & \textbf{6.7} & \textbf{5} \\
        LLM Modulo [Common]$_{\rm GPT-3.5-Turbo}$ & 100 & 67.9 & 16.7 & 14 & 5 & 2.8 \\

        LLM Modulo [Hard]$_{\rm GPT-3.5-Turbo}$ & 100 & 61.3 & 4.4 & 10.7 & 5.6 & 1.6 \\
        LLM Modulo [Json]$_{\rm GPT-3.5-Turbo}$ & 100 & 61.3 & 4.4 & 10.2 & 3.9 & 1.1 \\
        \textbf{LLM Modulo [All]$_{\rm GPT-4-Turbo}$} & \textbf{100} & \textbf{89.2} & \textbf{40.6} & \textbf{62.1} & \textbf{39.4} & \textbf{20.6} \\
        \bottomrule
    \end{tabular}
    \caption{
    We report the results on TravelPlanning Validation set following \cite{xie2024travelplanner}. Grayed out results on CoT / ReAct / Reflexion variants are reported from \cite{xie2024travelplanner} for completeness. Direct$_{\text{Model}}$ are reproduced baselines as implemented in \cite{xie2024travelplanner}. $\text{LLM Modulo}[\text{Crtic}]_{\mathcal{M}}$ represents the critics used during the LLM-Modulo planning with model $\mathcal{M}$. Values are percentages of delivery rate, micro and macro commonsense and hard constraints and finally, success rate defined as \textit{Final Pass Rate} as in \cite{xie2024travelplanner}.
    }
    \label{tab:model_performance}
\end{table*}

Our implementation follows the LLM-Modulo architecture presented in \cite{kambhampati2024llms} and the LLM-Modulo for TravelPlanning can be seen in Fig. \ref{fig:llm_modulo_TP}.

\textbf{Problem Specification} By design the TravelPlanning domain presents queries that contains all information which maybe required to generate a feasible travel plan, however, the query is in natural language which is a popular mode of interacting with LLMs. 

\textbf{Prompt Generator} Consistent with use of LLMs as agents \cite{wang2024survey,xi2023rise,chang2024survey}, we provide an instruction prompt the LLM along with the context information about flights, hotels etc. We also provide instructions on the output format of the generated plan and present few shot example. This is directly inherited from the implementation of \cite{xie2024travelplanner}.

\textbf{Plan Backboard and Reformatter} We transform the LLM generated natural language travel plan into a valid JSON format and store it in the plan blackboard. This translation is done through the use of LLM as a reformulator and we reuse it for our model based critics which require structured parseable plans. 

\textbf{Crtics} All of the critics that we use are binary critics paired with a backprompt describing the issue should the critic detect one. The Format critics ensures the syntactic validity of the plan such as validating the JSON and eliminating any missing key-values which is a precondition for all other critics, therefore takes precedence. We repurpose the common-sense constraints as style critics that provide information about missing implicit preference considerations and finally use the hard-constraints as the remainder of the critics.

\textbf{Metacontroller} All of the critics evaluate a generated plan and incase any of the critics find issues with the generated plan the metacontroller takes on the control flow. It contains the decision-making logic to stich together the critic responses, choose which backprompts to allow for (if a certain pedagogical prompting is in effect) or other consolidation of various backprompts. The metacontroller interfaces with the LLM and makes use of the Prompt Generator to contain other information such as instructions, database context, formatting and few shot examples along with the compiled backprompt. In this work we concatenate the backprompts from all the critics and add it to the initial prompt and provide it to the LLM.  


The interaction loop in LLM Modulo continues uptill a specificed maximum budget (set to 10 iterations) or until all of the critics agree to the generated plan. Building on this integration, the use of critics within the Modulo framework illustrates that similar evaluative mechanisms can be effectively utilized across different datasets by converting traditional evaluation constraints into critics, enhancing output precision and adaptability. Moreover, the employment of a rudimentary \textit{metacontroller} highlights the substantial potential for advancement. The current approach, which aggregates and reiterates critic responses, is simple yet effective. Future enhancements could include strategically ordering constraints or providing more targeted and relevant feedback in critic responses, improving the system’s efficacy and responsiveness.

\section{Experiments and Results}
\vspace{-0.2cm}

The baseline results with GPT-3.5 Turbo model showed a final pass rate of 0\%, for both micro and macro pass rates in commonsense and hard constraints being low across 180 queries, indicating that none of the generated plans fully met all constraints. Surprisingly, methods such as Chain of Thought\cite{wei2022chain}, ReAct \cite{yao2022react} and Reflexion\cite{shinn2024reflexion} provides no improvement. When used with GPT-3.5 Turbo, Chain of Thought and Reflexion exhibit a final pass rate of 0\%, while Chain of Thought alone achieves a slightly higher pass rate of 0.6\%, indicating suboptimal performance. While improvements from prompt engineering often remain unexplained, the LLM-Modulo framework promises soundness of produced plans consistent with the critics used. Indeed, we find that our LLM Modulo planner with GPT3.5-Turbo (older model) surpasses GPT4-Turbo baseline performance (newer model). Consequently we see improved micro/macro pass rates along common-sense and hard constraints. LLM Modulo GPT4-Turbo achieves state of the art performance on TravelPlanning benchmark under the agentic LLM paradigm of using LLMs to generate final plans. We achieve 20.6\% final pass rate compared to 4.4\% baseline. Moreover, such gains are well-founded and the source of improvement can be attributed to the presence of reliable critics in the planning pipeline.

\subsection{Ablations}
\vspace{-0.2cm}

We categorized the critics into three subgroups: Format (which includes checks for valid JSON and the presence of all key-value pairs), Hard (hard constraints), and Commonsense (commonsense constraints). We study the impact of each class of critics on the final performance and other fine-grained metrics such as micro/macro rates. Note that choosing a subset of critics implies that we prevent the LLM from getting pointed feedback on issues in the generated plan as well as allow for suboptimal plans (in that they are guarateed to satisfy only the subset of critics) as final result.

As anticipated, LLM Modulo with a subset of critics underperforms relative to the model that uses all critic types, yet they demonstrated improvements over baselines ($\text{Direct}_{\text{Model}}$) and CoT/ReAct/Reflexion variants. Utilizing solely the commonsense critics resulted in a final pass rate of 2.8\%, while employing just the hard constraints as critics achieved a final pass rate of 1.6\%. Solely ensuring the correct format yielded a 1.1\% final pass rate.

We note our results demonstrate the composability of the critics. Compared to [Hard], [Common], or [JSON] variants, the LLM Modulo [All] result is higher micro/macro pass rates across common-sense and hard constraints (with exceptions for delivery rate and common-sense micro pass rate). While providing just the Commonsense critics gave the most improvement, composing it with other critics (Hard, JSON) yields a much higher performance rate. Finally, we see that even thought $\text{Direct}_{\text{GPT-4-Turbo}}$ has higher micro/macro pass rates even comparable to $\text{LLM Modulo[All]}_{\text{GPT-4-Turbo}}$, the final pass rate in the case of LLM Modulo is ~ 4.6x higher.

\subsection{Frequency analysis of Critics}
\vspace{-0.2cm}

\begin{figure}[ht]
    \centering
    \includegraphics[width=0.9\linewidth]{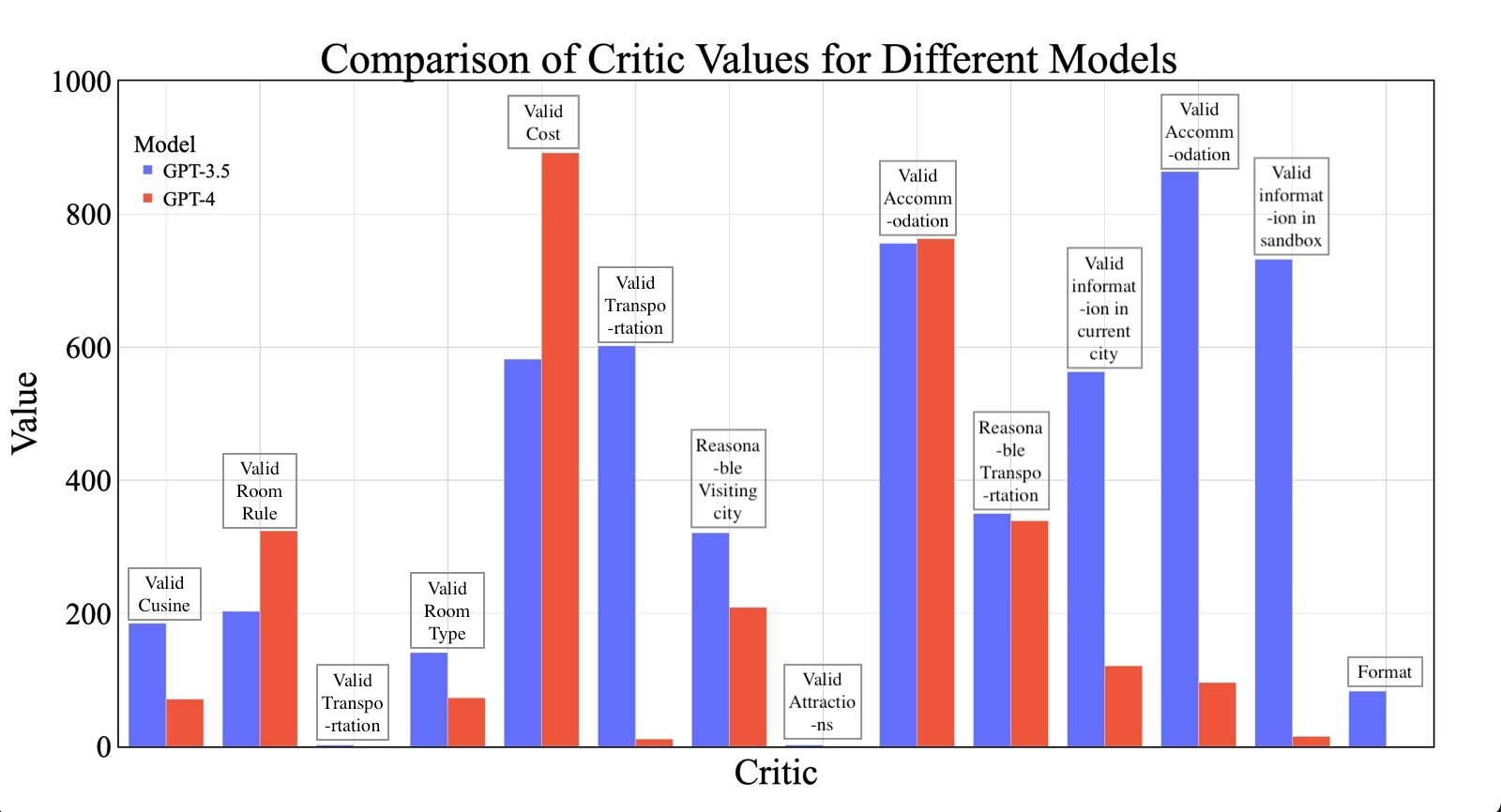}
    \caption{Comparison of Critic Values for GPT 3.5 Turbo and GPT 4 Turbo}
    \label{fig:critic_freq_gpt}
\end{figure}

LLMs are known for capturing common-sense information about real world tasks \cite{stechly2024chain,stechly2024selfverification,guan2024task,kambhampati2024can,Verma_2024}. Figure \ref{fig:critic_freq_gpt} shows the number of times a critic was fired (or detected an issue with a generated plan) across all 180 validation set instances and iteration steps in LLM Modulo for GPT3.5-Turbo and GPT4-Turbo models. We find that certain critics more frequent than others since the corresponding issues occur more often. We also find that the format critic (for ensuring JSON correctness) is required more often by GPT3.5-Turbo over GPT4-Turbo and that the LLM Modulo planner is able to resolve format issues in the first few iterations of the budget. The critics that often disagree with the plan are \textit{valid\_cost}, \textit{is\_valid\_accommodation}, and \textit{is\_valid\_information}, which are generated respectively by \textit{Budget}, \textit{Room Type}, and \textit{Validate Itinerary}.
With the knowledge that critics maybe correlated (such as change in accommodation impacts transport and budget) and that only a few critics are the flagged most of the times, future work may take such statistics into account when designing an advanced Metacontroller and identification of points of failure for LLM generated plans.

\subsection{Recovering critics from LLMs}
\vspace{-0.2cm}
Previous subsection highlights that only a few critics are flagged most of the times during the LLM Modulo interaction. We argue that LLMs may indeed be useful for extracting the implementation for such critics. This is akin to teasing out the model based critics in the LLM-Modulo frameworks. We prompt the GPT-4-Turbo model to obtain the implementation of the critics by providing it contextual information such as the objective of the critic, available tools or databases with corresponding function declaration (such as flights, etc.) and the input plan as JSON (along with the JSON schema). We do so for common-sense and hard constraint critics. We then compared the generated critics code implementation with the existing ones to evaluate their correctness. Typically, we observe that only minimal modifications were necessary (such as fixing function call signature and syntax which itself can be automated via critics such as compilers and parsers) for the generated critics to match the efficacy of the pre-existing ones. The generated hard critics included: \textit{Room Type}, \textit{Cuisines}, \textit{Budget}, and \textit{Transportation}. The generated commonsense critics encompassed: \textit{Complete Information}, \textit{Diverse Restaurants}, \textit{Diverse Attractions}, and \textit{Validate Itinerary}.


\begin{figure}[ht]
    \centering
    \includegraphics[width=0.9\linewidth]{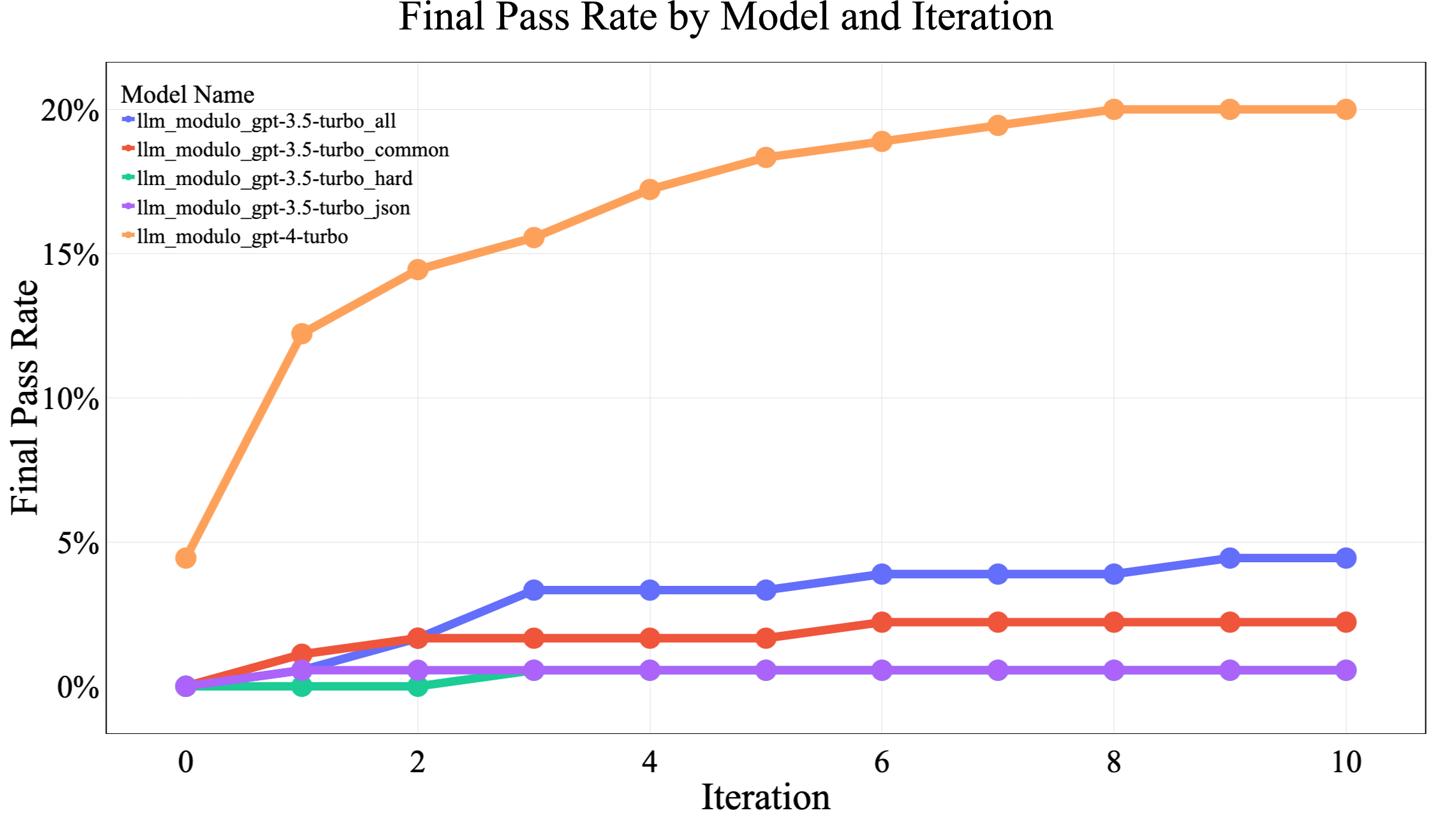}
    \caption{Final Pass rates of models across LLM Modulo Iterations}
    \label{fig:final_pass_rate_iter}
\end{figure}

\section{Conclusion}
\vspace{-0.2cm}

We demonstrate the effective application of the LLM modulo framework within the TravelPlanning domain showcasing a remarkable ~ 4.6x improvement performance for GPT4-Turbo achieveing new state of the art on TravelPlanning domain under the agentic LLM paradigm. Our work also validates the framework's robustness in real-world scenarios such as Travel Planning as motivated by \cite{xie2024travelplanner}. We showcase that such performance boost is well-founded and easily surpasses predominant ways of enhacing agentic abilities of LLMs such as CoT / ReAct and Reflexion. We do so by allowing critics to be part of the LLM-Modulo based planning pipeline. We also showcase that such critics may also be extracted through the LLMs (LLMs working towards teasing out model-based critics / verifiers). Finally, we showcase the LLMs use as a reformualor to translate natural language plans to a structured representation (JSON) that can be easily used by the critics. Along our discussion, we also point out potential next steps beyond our first investigation to further improve agentic LLM performance.

\bibliographystyle{icml2024}
\bibliography{example_paper}

\begin{thebibliography}{21}
\providecommand{\natexlab}[1]{#1}
\providecommand{\url}[1]{\texttt{#1}}
\expandafter\ifx\csname urlstyle\endcsname\relax
  \providecommand{\doi}[1]{doi: #1}\else
  \providecommand{\doi}{doi: \begingroup \urlstyle{rm}\Url}\fi

\bibitem[Chang et~al.(2024)Chang, Wang, Wang, Wu, Yang, Zhu, Chen, Yi, Wang, Wang, et~al.]{chang2024survey}
Chang, Y., Wang, X., Wang, J., Wu, Y., Yang, L., Zhu, K., Chen, H., Yi, X., Wang, C., Wang, Y., et~al.
\newblock A survey on evaluation of large language models.
\newblock \emph{ACM Transactions on Intelligent Systems and Technology}, 15\penalty0 (3):\penalty0 1--45, 2024.

\bibitem[Guan et~al.(2024)Guan, Zhou, Liu, Zha, Amor, and Kambhampati]{guan2024task}
Guan, L., Zhou, Y., Liu, D., Zha, Y., Amor, H.~B., and Kambhampati, S.
\newblock "task success" is not enough: Investigating the use of video-language models as behavior critics for catching undesirable agent behaviors, 2024.

\bibitem[Hsieh et~al.(2023)Hsieh, Chen, Li, Fujii, Ratner, Lee, Krishna, and Pfister]{hsieh2023tool}
Hsieh, C.-Y., Chen, S.-A., Li, C.-L., Fujii, Y., Ratner, A., Lee, C.-Y., Krishna, R., and Pfister, T.
\newblock Tool documentation enables zero-shot tool-usage with large language models.
\newblock \emph{arXiv preprint arXiv:2308.00675}, 2023.

\bibitem[Kahneman(2011)]{kahneman2011thinking}
Kahneman, D.
\newblock \emph{Thinking, fast and slow}.
\newblock macmillan, 2011.

\bibitem[Kambhampati(2024)]{kambhampati2024can}
Kambhampati, S.
\newblock Can large language models reason and plan?
\newblock \emph{Annals of the New York Academy of Sciences}, 1534\penalty0 (1):\penalty0 15--18, 2024.

\bibitem[Kambhampati et~al.(2024)Kambhampati, Valmeekam, Guan, Stechly, Verma, Bhambri, Saldyt, and Murthy]{kambhampati2024llms}
Kambhampati, S., Valmeekam, K., Guan, L., Stechly, K., Verma, M., Bhambri, S., Saldyt, L., and Murthy, A.
\newblock Llms can't plan, but can help planning in llm-modulo frameworks, 2024.

\bibitem[OpenAI(2022)]{openai2022gpt35}
OpenAI.
\newblock Gpt-3.5: Language model, 2022.
\newblock \url{https://www.openai.com}.

\bibitem[OpenAI(2023)]{openai2023gpt4}
OpenAI.
\newblock Gpt-4: Language model, 2023.
\newblock \url{https://www.openai.com}.

\bibitem[Paranjape et~al.(2023)Paranjape, Lundberg, Singh, Hajishirzi, Zettlemoyer, and Ribeiro]{paranjape2023art}
Paranjape, B., Lundberg, S., Singh, S., Hajishirzi, H., Zettlemoyer, L., and Ribeiro, M.~T.
\newblock Art: Automatic multi-step reasoning and tool-use for large language models.
\newblock \emph{arXiv preprint arXiv:2303.09014}, 2023.

\bibitem[Schick et~al.(2024)Schick, Dwivedi-Yu, Dess{\`\i}, Raileanu, Lomeli, Hambro, Zettlemoyer, Cancedda, and Scialom]{schick2024toolformer}
Schick, T., Dwivedi-Yu, J., Dess{\`\i}, R., Raileanu, R., Lomeli, M., Hambro, E., Zettlemoyer, L., Cancedda, N., and Scialom, T.
\newblock Toolformer: Language models can teach themselves to use tools.
\newblock \emph{Advances in Neural Information Processing Systems}, 36, 2024.

\bibitem[Shinn et~al.(2024)Shinn, Cassano, Gopinath, Narasimhan, and Yao]{shinn2024reflexion}
Shinn, N., Cassano, F., Gopinath, A., Narasimhan, K., and Yao, S.
\newblock Reflexion: Language agents with verbal reinforcement learning.
\newblock \emph{Advances in Neural Information Processing Systems}, 36, 2024.

\bibitem[Stechly et~al.(2024{\natexlab{a}})Stechly, Valmeekam, and Kambhampati]{stechly2024chain}
Stechly, K., Valmeekam, K., and Kambhampati, S.
\newblock Chain of thoughtlessness: An analysis of cot in planning, 2024{\natexlab{a}}.

\bibitem[Stechly et~al.(2024{\natexlab{b}})Stechly, Valmeekam, and Kambhampati]{stechly2024selfverification}
Stechly, K., Valmeekam, K., and Kambhampati, S.
\newblock On the self-verification limitations of large language models on reasoning and planning tasks, 2024{\natexlab{b}}.

\bibitem[Valmeekam et~al.(2022)Valmeekam, Olmo, Sreedharan, and Kambhampati]{valmeekam2022large}
Valmeekam, K., Olmo, A., Sreedharan, S., and Kambhampati, S.
\newblock Large language models still can't plan (a benchmark for llms on planning and reasoning about change).
\newblock \emph{arXiv preprint arXiv:2206.10498}, 2022.

\bibitem[Verma et~al.(2024{\natexlab{a}})Verma, Bhambri, and Kambhampati]{Verma_2024}
Verma, M., Bhambri, S., and Kambhampati, S.
\newblock Theory of mind abilities of large language models in human-robot interaction: An illusion?
\newblock In \emph{Companion of the 2024 ACM/IEEE International Conference on Human-Robot Interaction}, HRI ’24. ACM, March 2024{\natexlab{a}}.
\newblock \doi{10.1145/3610978.3640767}.
\newblock URL \url{http://dx.doi.org/10.1145/3610978.3640767}.

\bibitem[Verma et~al.(2024{\natexlab{b}})Verma, Bhambri, and Kambhampati]{verma2024theory}
Verma, M., Bhambri, S., and Kambhampati, S.
\newblock Theory of mind abilities of large language models in human-robot interaction: An illusion?
\newblock In \emph{Companion of the 2024 ACM/IEEE International Conference on Human-Robot Interaction}, pp.\  36--45, 2024{\natexlab{b}}.

\bibitem[Wang et~al.(2024)Wang, Ma, Feng, Zhang, Yang, Zhang, Chen, Tang, Chen, Lin, et~al.]{wang2024survey}
Wang, L., Ma, C., Feng, X., Zhang, Z., Yang, H., Zhang, J., Chen, Z., Tang, J., Chen, X., Lin, Y., et~al.
\newblock A survey on large language model based autonomous agents.
\newblock \emph{Frontiers of Computer Science}, 18\penalty0 (6):\penalty0 1--26, 2024.

\bibitem[Wei et~al.(2022)Wei, Wang, Schuurmans, Bosma, Xia, Chi, Le, Zhou, et~al.]{wei2022chain}
Wei, J., Wang, X., Schuurmans, D., Bosma, M., Xia, F., Chi, E., Le, Q.~V., Zhou, D., et~al.
\newblock Chain-of-thought prompting elicits reasoning in large language models.
\newblock \emph{Advances in neural information processing systems}, 35:\penalty0 24824--24837, 2022.

\bibitem[Xi et~al.(2023)Xi, Chen, Guo, He, Ding, Hong, Zhang, Wang, Jin, Zhou, et~al.]{xi2023rise}
Xi, Z., Chen, W., Guo, X., He, W., Ding, Y., Hong, B., Zhang, M., Wang, J., Jin, S., Zhou, E., et~al.
\newblock The rise and potential of large language model based agents: A survey.
\newblock \emph{arXiv preprint arXiv:2309.07864}, 2023.

\bibitem[Xie et~al.(2024)Xie, Zhang, Chen, Zhu, Lou, Tian, Xiao, and Su]{xie2024travelplanner}
Xie, J., Zhang, K., Chen, J., Zhu, T., Lou, R., Tian, Y., Xiao, Y., and Su, Y.
\newblock Travelplanner: A benchmark for real-world planning with language agents.
\newblock \emph{arXiv preprint arXiv:2402.01622}, 2024.

\bibitem[Yao et~al.(2022)Yao, Zhao, Yu, Du, Shafran, Narasimhan, and Cao]{yao2022react}
Yao, S., Zhao, J., Yu, D., Du, N., Shafran, I., Narasimhan, K., and Cao, Y.
\newblock React: Synergizing reasoning and acting in language models.
\newblock \emph{arXiv preprint arXiv:2210.03629}, 2022.

\end{thebibliography}

\newpage
\appendix
\onecolumn

\section{Resources Used}
In this work we leverage OpenAI API for prompting the Language Models. We use \texttt{gpt-4-0613} for GPT4, \texttt{gpt-3.5-turbo-0125}, for all our experimentation in April-May 2024.

\section{Extracted Critic}
A substantial number of critics could be derived from the foundational LLM. These include frequently mentioned aspects as depicted in results \ref{fig:critic_freq_gpt}, such as "Room Type," "Cuisines," "Budget," "Transportation," "Complete Information," "Diverse Restaurants," and "Diverse Attractions." An example is illustrated below:
"\textbf{Budget}"
Prompt:

\begin{numberedbox}[label={  prompt:rq3-Domain  }]{ Extracting Hard Constraint : Budget }
{\scriptsize
\begin{verbatim}
    Assume you have a json with following as a list. The list is the itinerary generated for each day of the 
    
    trip. 

```
- llm\_response: An array containing objects with the following fields for each day of the trip:
  - day: Integer representing the day number.
  - people\_number: Integer representing the number of people.
  - current\_city: String representing the current city/location.
  - transportation: String representing transportation details (e.g., flight number, departure/arrival time).
  - breakfast: String representing breakfast arrangements.
  - attraction: String representing the attraction(s) for the day.
  - lunch: String representing lunch arrangements.
  - dinner: String representing dinner arrangements.
  - accommodation: String representing accommodation details.
```

Additionally, you have following functions that you can use : 

1. get\_cost\_of\_transport(source, destination, mode-of-travel)['cost']
2. A valid restaurants dataframe with keys : 'Name','Average Cost','Cuisines','Aggregate Rating','City'
3. A valid attractions dataframe with keys : 'Name','Latitude','Longitude','Address','Phone','Website',"City"
4. A valid accomodations dataframe with keys : 'NAME','price','room type', '

house_rules', 'minimum nights', 'maximum occupancy', 'review rate number', 'city'
5. A valid flights dataframe with kyes : 'Flight Number', 'Price', 'DepTime', 'ArrTime', 

'ActualElapsedTime','FlightDate','OriginCityName','DestCityName','Distance'


You task is to write 

A method called 'get\_total\_cost' which parses the json file and calculates the total cost of the trip. 

Ensure you take into account the cost of transport, breakfast, lunch, dinner, accomodation.

Ensure that you take into account the number of people for which the itinerary has been made. 

Use the data within the json and the available dataframes and tools for estimating the total cost. 

Do not assume any other details.

Implement the complete function with all the details. Return a float value representing the total cost. 

\end{verbatim}

}
\end{numberedbox}

\newpage
\section{Additional Analysis}

\begin{figure}[h]
    \centering
    \includegraphics[width=0.7\linewidth]{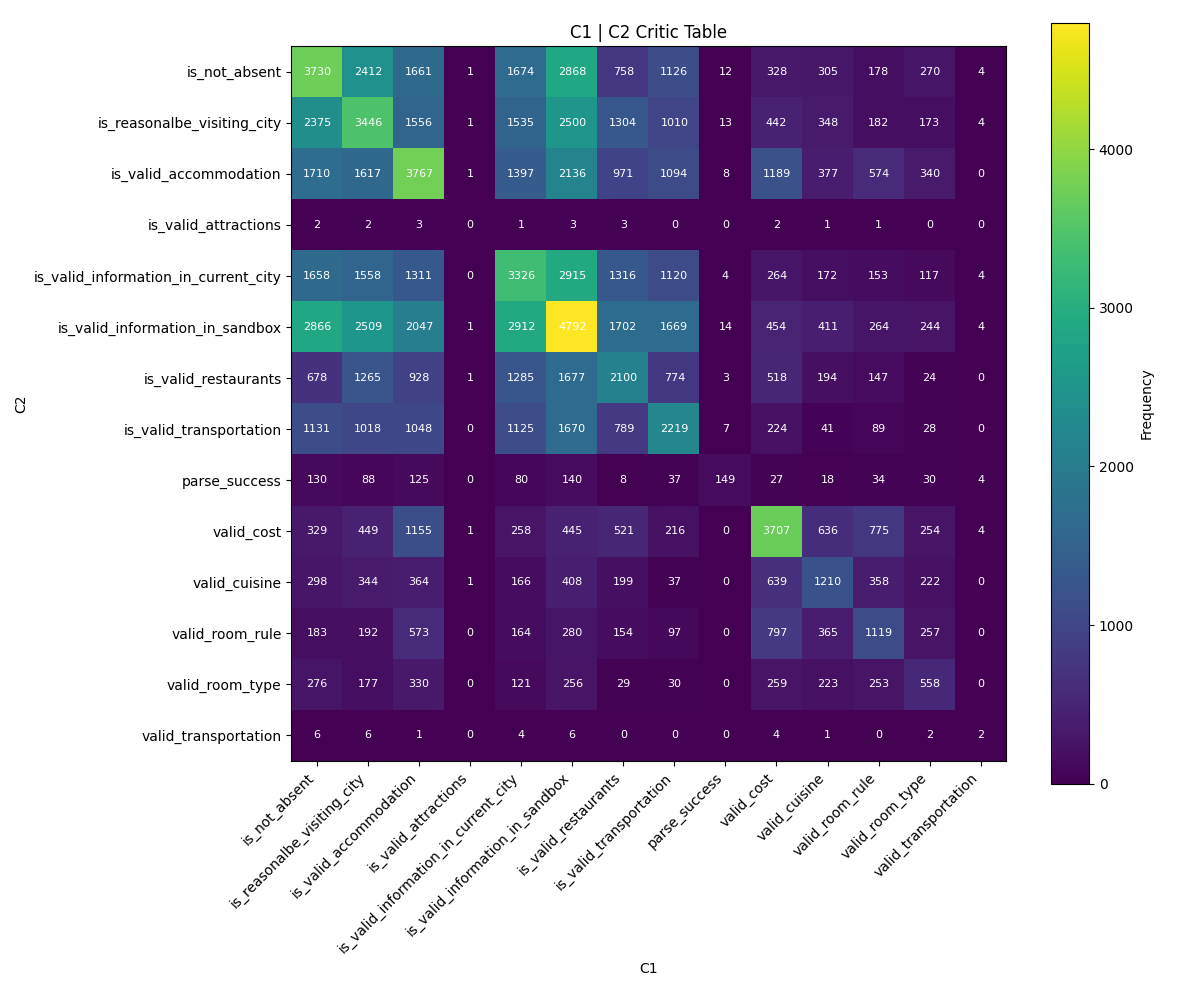}
    \caption{Enter Caption}
    \label{fig:C1___C2_Critic_Table}
\end{figure}

We analyze the correlation between, given that a certain critic was fired (C1) how many times another critic was fired (C2). This is an exploratory analsis to find the correlation between how critics were fired and obtain information about failures during the LLM Modulo iteration steps. We leave a thorough analysis as future work. In figure \ref{fig:C1___C2_Critic_Table} we see that several critics are correlated and failure of one almost always triggers a certain other critic.

\section{Recovering Test set Fields}

Generating plans for the queries in the test set included an additional step. The test set does not include essential features like budget, people\_number, local\_constraints, and days which are essential to the critics to evaluate a plan generated by an LLM. To address this gap, we utilized GPT-4 Turbo to extract these features. By passing the user query as the prompt, we instructed the model to extract all necessary features in the required format.

\end{document}